# Rapid online learning and robust recall in a neuromorphic olfactory circuit

Nabil Imam[1*], Thomas A. Cleland[2*]

[1] Neuromorphic Computing Laboratory, Intel Corporation, San Francisco, CA 94111, USA. nabil.imam@intel.com.

[2] Computational Physiology Laboratory, Dept. Psychology, Cornell University, Ithaca, NY 14853, USA.  tac29@cornell.edu.

* Correspondence to either author.

**Abstract**

We present a neural algorithm for the rapid online learning and identification of odorant samples under noise, based on the architecture of the mammalian olfactory bulb and implemented on the Intel Loihi neuromorphic system. As with biological olfaction, the spike timing-based algorithm utilizes distributed, event-driven computations and rapid (one-shot) online learning. Spike timing-dependent plasticity rules operate iteratively over sequential gamma-frequency packets to construct odor representations from the activity of chemosensor arrays mounted in a wind tunnel. Learned odorants then are reliably identified despite strong destructive interference. Noise resistance is further enhanced by neuromodulation and contextual priming. Lifelong learning capabilities are enabled by adult neurogenesis. The algorithm is applicable to any signal identification problem in which high-dimensional signals are embedded in unknown backgrounds.

Spike timing-based mechanisms of coding and computation operating within plastic neural circuits present a central problem of interest to both neuroscience and neuromorphic computing. We have found that a coordinated set of these mechanisms, hypothesized for the neural circuitry of the external plexiform layer (EPL) of the mammalian main olfactory bulb (MOB), exhibits rapid learning of arbitrary high-dimensional neural representations and robust memory recall despite occlusion by random sources of destructive interference. Based on these mechanisms, we derived a neural algorithm for the learning of odorant signals and their robust identification under noise, and instantiated it in the Intel Loihi neuromorphic system [1]. The algorithm operates over a network of excitatory and inhibitory units that embed feedforward and recurrent feedback circuit motifs. Information in the network is represented by sparse patterns of spike timing [2] measured against an underlying network rhythm. Learning is based on local spike timing-dependent plasticity rules, and memory is retrieved over sequential gamma-breadth packets of spiking activity. The network can be effectively trained using one-shot learning, and innately supports online learning; that is, additional training on new stimuli does not disrupt prior learning.

While both biological and artificial olfaction systems recognize chemical analytes based on activity patterns across arrays of weakly specific chemosensors [3,4], mammalian olfaction demonstrates levels of performance in signal restoration and identification currently unmatched by artificial systems. Indeed, the underlying identification problem is deceptively difficult. Natural odors comprise mixtures of many different odorant molecules [5]; moreover, under natural conditions, different odors from many separate sources intermingle freely and, when sampled together, chemically occlude one another in competition for primary chemosensor binding sites [6-9]. This occlusion substantially disrupts the primary sensory activation patterns that provide the basis for odor recognition. Moreover, the patterns of potential occlusion are unrelated to the input statistics of the odors of interest, and hence unpredictable. This presents an extraordinary signal restoration challenge that has been recognized as one of the central problems in neuromorphic olfaction [3, 10, 11]. By designing a neuromorphic algorithm based on computational principles extracted from the biological system, and implementing it on a compact, field-deployable hardware platform, we sought to dramatically improve the performance and capabilities of artificial chemosensory systems deployed into uncontrolled environments.

This biological system exhibits several critical properties and mechanisms that we used to address the problem. Primary sensory representations of odor stimuli at steady state constitute intrinsically high-dimensional feature vectors, the dimensionality of which is defined by the number of receptor types (columns) expressed by the olfactory system [12] ; this number ranges from the hundreds to over 1000 in different mammalian species. Each of these receptor types induces spiking in a corresponding group of principal neurons (mitral cells; MCs). Mechanistically, fast coherent oscillations in the gamma band (c. 30-80 Hz), which are intrinsic to MOB circuitry [13-15], phase-restrict the timing of these MC action potentials [13, 16]. This property discretizes spiking output into gamma-breadth packets, here enabling a robust within-packet phase precedence code [17, 18] that disambiguates phase-leading from phase-lagging spikes within the permissive epoch of each gamma cycle. Recurrent activity loops in OB circuitry evince control systems architecture, implementing gain control in the superficial layers [19-21] and enabling autoassociative attractor dynamics in the deeper network [22]. Odor learning in the biological system is localized and rapid, and depends substantially on plastic synapses within the MOB [23-27, 28, 29], here instantiated as spike timing-dependent plasticity rules. The neuromodulatory tuning of MOB circuit properties [30-32] here is leveraged as an optimization trajectory rather than a fixed state variable. Adult neurogenesis in the MOB, known to be required for odor learning and memory [26, 27, 33, 34], here provides indefinite capacity for lifelong learning through the permanent differentiation and replacement of plastic interneurons.

Our algorithm is derived from these computational properties of the EPL neural circuit in the biological MOB. We train and test the algorithm using data from the Vergara et al. dataset [4], acquired from an array of 72 chemosensors mounted across a wind tunnel, and show that it rapidly learns odor representations and robustly identifies learned odors under high levels of destructive interference, as well as in the presence of natural variance arising from odorant plume dynamics. The destructive interference model, impulse noise, is designed to model the effects of intermixed, simultaneously sampled background odorants that effectively randomize the activation levels of a substantial fraction of the primary sensors on which odor recognition depends. The algorithm exhibits online learning and generalizes broadly beyond experience; accordingly, it can be trained on relatively clean diagnostic samples using one- or few-shot learning and then deployed into environments containing unknown contaminants and other sources of interference.

**Results**

*Model architecture*

The structure of the model network (Figure 1a) was based on the circuitry and computational properties of the mammalian MOB, optimized for efficient implementation as a spiking network on the Loihi chip (Figure 1b). In particular, we instantiated some core principles of MOB computation that we have hypothesized for the biological system [12, 14, 15, 17, 35], including (1) the dynamically acquired, learning-dependent topology of the lateral inhibitory network of the EPL,

(2) the importance of gamma-discretized spike timing-based computation in the EPL, (3) the principle that MCs deliver excitation to granule cell interneurons (GCs) irrespective of distance, whereas GCs effectively inhibit MCs only locally, and only via GC spiking, (4) the principle that this inhibition of MCs by GCs predominantly manifests as delays in MC spike times on the gamma timescale, (5) the principle that these fine-timescale EPL computations do not meaningfully influence the coarse-timescale computations of the glomerular layer, (6) the principle that only a minority of principal neurons participate in gamma dynamics during any given stimulus presentation, (7) the permanent differentiation of GCs by the process of odor learning, and the consequent need for replacement by adult neurogenesis, and (8) the utility of treating neuromodulation as an optimization trajectory rather than as a stationary state.

Like the mammalian MOB, the neuromorphic EPL network is implicitly columnar (Figure 1a). Each column comprises a single MC principal neuron as well as up to 50 inhibitory GC interneurons, coupled by moderately sparse intercolumnar excitatory synapses (connection probability = 0.2) and local (intracolumnar) inhibitory synapses (see *Methods*). We activated the MCs of a 72-column EPL network using the "Gas sensor arrays in open sampling settings" dataset published by Vergara and colleagues [4] and available from the UCI Machine Learning Repository (https://bit.ly/2nteY78). Samples were drawn from an array of 72 metal oxide gas sensor (MOS) elements spatially distributed across the full 1.2 m breadth of a wind tunnel [4] (Figure 1c). From the 180 s datastreams comprising each odorant presentation in this dataset, sensor array responses were sampled ("sniffed") from a common point in time and presented to the EPL network for training or testing. That is, individual odor samples ("sniffs") comprised discrete feature vectors in which the pattern of amplitudes across vector elements reflected odor quality, as well as concentration-based variance owing to plume dynamics in the wind tunnel.

The biological EPL network is intrinsically oscillogenic in the gamma band (30-80 Hz) [15, 36], and MC action potentials are statistically phase-constrained with respect to these local oscillations [13, 16]. In our algorithm, MC spikes were constrained in time by an ongoing network oscillation with alternating permissive and inhibitory epochs reflecting the periodic inhibition of the OB gamma cycle [14, 36] (Figure 1d). Sensory integration and MC spiking were enabled only during permissive epochs, whereas inhibitory epochs reset and held the activation of all MCs at zero. Therefore, in the absence of learning, and given stationary sensor input, the temporal patterning of spikes evoked by a given odor directly reflected sensor activation levels – stronger excitation evoked correspondingly earlier spikes – and was repeated across successive gamma cycles (Figure 1d). Different odors evoked correspondingly different spatiotemporal spike patterns across the MC population, thereby generating a hybrid channel/phase code, or *precedence code*, on the gamma timescale.

Critically, this dynamical architecture enables multiple iterative cycles of processing for each sample by taking advantage of the differences in timescale between sampling (4-8 Hz in rodent sniffing, 100 Hz in the Vergara et al. dataset) and processing (30-80 Hz gamma oscillations in the rodent olfactory bulb, 100 kHz in the Loihi chip). In the present instantiation of the algorithm, five gamma cycles, each requiring 0.4 ms to execute (see *Methods*), were embedded within each odor presentation ("sniff") for both training and testing. After learning, GC feedback inhibition on each successive gamma cycle iteratively modified MC spike timing and hence altered the precedence code. Network output thus was interpreted as an evolving series of representations, in which each discrete representation comprised a population of spikes, with each spike defined by the identity of the active MC and the spike latency within the corresponding gamma cycle. These representations then were classified based on their similarities to each of the representations known by the network.

*Excitatory plasticity determines GC receptive fields*

Each gamma-constrained array of MC action potentials, in addition to serving as network output, also drove its complement of postsynaptic GCs across the network. During learning, the synaptic weights between MCs and GCs were systematically modified by experience. GCs were modeled as single-compartment neurons that accumulated excitatory synaptic inputs from their widely distributed presynaptic MCs. Upon reaching threshold, they generated spike events that inhibited their cocolumnar postsynaptic MC in the subsequent gamma cycle.

GC spiking also initiated excitatory synaptic plasticity. Specifically, GCs learned to respond to higher-order stimulus features by becoming selective for specific combinations of MC spiking activity. To do this, we implemented a spike timing-dependent plasticity (STDP) rule that learned these input combinations in terms of a spike phase precedence coding metric on the gamma timescale [17]. Under initial conditions, GCs required moderately synchronous spike inputs from several presynaptic MCs in order to evoke an action potential. Classical STDP most powerfully strengthens the synaptic weights of synapses mediating presynaptic spikes that immediately precede a postsynaptic spike; we implemented this principle with a heterosynaptic additive STDP rule that strengthened these synapses and weakened all other incoming synapses, including those in which the presynaptic MC spiked at other times or not at all (Figure 2a). Accordingly, spiking GCs ultimately learned a fixed dependency on the synchronous firing of a set of $k$ MC inputs, with inputs from other MCs decaying to zero (effectively a "$k$ winners take all" learning rule). Consequently, at the end of the training period, the response to each trained odorant evoked a distributed ensemble of GCs tuned to a diversity of stimulus-specific higher-order correlation patterns (Figure 2b).

*Inhibitory plasticity denoises MC representations*

Spikes evoked by GC interneurons delivered synaptic inhibition onto the MC of their local column. As proposed for the biological system, the weights of GC-mediated inhibitory inputs regulated the timing of MC spikes within the permissive phase of the gamma cycle, with stronger weights imposing greater MC spike time delays within each gamma cycle [14, 35, 37]. In the neuromorphic system, a GC spike blocked the generation of a spike on its follower MC for a period of time corresponding to the inhibitory synaptic weight. During odor learning, the durations of GC spike-evoked inhibitory windows were iteratively modified until the release of inhibition on the MC soma coincided with a threshold crossing in the MC apical dendrite resulting from integrated sensory input (Figure 2c). During testing, the end of the GC inhibitory window permitted the MC to fire, and evoked a rebound spike in the MC even in the absence of sufficient apical dendritic input. Synaptic inputs from multiple local GCs onto a common MC were independent of one another, enabling a diverse range of higher-order GC receptive fields to independently affect the MC. During testing, occluded inputs activated some fraction of GCs, which then modified their postsynaptic MC spike times such that the representation in the next gamma cycle was closer to a learned odorant, hence activating an increased fraction of its corresponding GCs. This process continued iteratively until the learned representation was recalled (Figure 2d).

This inhibitory plasticity rule enables the EPL network to learn the timing relationships among MC spikes in response to a given odor stimulus. Consequently, because relative spike times signify MC activation levels, the network effectively learns the specific ratiometric pattern of activation levels among MCs that characterizes a given odor. This spatiotemporal basis for odor representation enables a substantially greater memory capacity than would be possible with spatial patterning alone; for example, two odors that activate the same population of MCs, but at different relative levels, can readily be distinguished by the trained network. Moreover, it consumes fewer spikes than rate-coding metrics, and can be read out much more quickly because it does not need to integrate multiple spikes over time to estimate rate. Finally, this spike timing-based metric for relational encoding, coupled with odor-specific profiles of feedback inhibition, renders these memory states as attractors, enabling incoming stimuli to be correctly classified by the trained network despite surprisingly high degrees of destructive interference. The trained EPL network thus comprises a spike-timing based autoassociator, embedding an arbitrary number of content-addressable memories.

*Odor learning enables identification of occluded stimuli*

We first trained the 72-column network on the odorant toluene in one shot (i.e., one sniff, enabling learning over five gamma cycles), and then, with plasticity disabled, tested the response of the trained network to presentations of toluene contaminated with destructive interference. To generate this interference, we entirely replaced a proportion $P$ of the sensory inputs with random values (*impulse noise*, $P = 0.6$ unless otherwise indicated) to represent strong and unpredictable receptor occlusion through simultaneous activation or inhibition by other ambient odorants. The occluded inputs remained consistent over the five gamma cycles of a sniff. In a naïve network, the presentation of occluded toluene yielded an essentially stationary and uninformative representation (Figure 3a, d). However, in the trained network, the spiking activity generated by occluded toluene was attracted over the five gamma cycles toward the previously learned toluene representation, enabling clear identification of the occluded unknown (Figure 3b, d-e). In contrast, if inhibitory plasticity (Figure 2c-d) was suppressed during training, the trained EPL network was unable to denoise the MC representation (Figure 3c-d).

As hypothesized for the biological olfactory bulb, odor learning in the network induces the permanent differentiation of granule cells (Figure 3f) that thereby become selective for higher-order feature combinations that are relatively diagnostic of the learned odor [38-40] (Figure 2b). We tested whether increased allocations of GCs, enabling each MC to be inhibited by a broader selection of feature combinations, would improve odor learning and identification under noise. We found that increasing the number of undifferentiated GCs per column improved the robustness of signal restoration, increasing the similarity of the occluded signal to the learned representation after five gamma cycles (Figure 3g). Nevertheless, we limited our simulations to five GCs per trained odorant and five gamma cycles per sniff in order to avoid ceiling effects and thereby better reveal the variables of greatest interest.

*Adult neurogenesis enables lifelong learning*

This learning algorithm irreversibly consumes GCs. Each odor memory is associated with a distributed population of differentiated GCs tuned to its complex diagnostic features. Fully differentiated, mature GCs do not undergo further plasticity and hence are protected from catastrophic interference [41, 42]. The learning of successively presented new odorants, however, would be increasingly handicapped by the declining pool of undifferentiated GCs (Figure 3g). The competition among distinct new odorants can be substantially reduced by sparser initial MC→GC connectivity and higher numbers of GCs, among other parameters [38]; however, genuine lifelong learning in such a system requires a steady source of undifferentiated GCs. Exactly this resource is provided to the mammalian olfactory system by constitutive adult neurogenesis. We propose that the critical role of adult neurogenesis in odor learning [26, 27, 33, 34, 43] be interpreted in this light.

In the neuromorphic algorithm, constitutive adult neurogenesis was simulated by configuring a new set of five GCs in every column after each successively learned odor stimulus. Hence, training a 72-column network on ten odors yielded a network with 3600 differentiated GCs. New GCs each received initial synaptic connections from a randomly selected 20% of the MCs across the network, and delivered inhibition onto their cocolumnar MC.

*Online learning of multiple representations*

We then trained the 72-column network sequentially with all ten odorants in the dataset [4], using a one-shot training regimen for each odor. In each case, the network was trained on one odor first, followed by a second odor, then by a third, until all ten odors had been learned. Similar results were obtained irrespective of the order in which the ten odorants were trained. A set of new, undifferentiated GCs was added to the network after each odor was learned, reflecting the effects of adult neurogenesis. Critically, subsequent odor training did not disrupt the memories of previously learned odors; that is, the EPL network supports robust online learning, and is resistant to catastrophic forgetting. This capacity for online learning is essential for memory formation under natural conditions, as well as for continuous device operation in the field; in either case, new signals of potential significance may be encountered at unpredictable times, and must be incorporated nondestructively into an existing knowledge base.

We then tested the algorithm's capacity to recognize and classify odorant samples that were strongly occluded by impulse noise, reflecting the effects of any number of independent odorous contaminants that could mask the odor of interest in uncontrolled environments. Following training on all ten odorants, sensor-evoked activity patterns generated by strongly occluded odor stimuli (impulse noise $P = 0.6$) were attracted specifically towards the learned representation of the corresponding odor (Figure 4a-c). Notably, the same network was able to rapidly identify occluded instances of all ten odors within five gamma cycles (Figure 4d). An odor was considered identified when the spatiotemporal pattern of its evoked spiking activity exceeded a Jaccard similarity [44] of 0.75 to one of the network's learned representations. Performance on this dataset under standard conditions was strong up to sample occlusion levels of $P = 0.6$, after which increased occlusion began to gradually impair classification performance (Figure 4e).

*Neuromodulation and cortical priming improve classification performance*

Neuromodulators like acetylcholine and noradrenaline generate powerful effects on stimulus representations and plasticity in multiple sensory systems including olfaction. Traditionally, they are treated as state variables that may sharpen representations, gate learning, or bias a network towards one source of input or another [45-47]. We instead modeled neuromodulatory effects as a dynamic search trajectory. Specifically, as the neuromodulator is released in response to active olfactory investigation (sampling), the local concentration around effector neurons and synapses rises over the course of successive sniffs, potentially enabling the most effective of the transient neuromodulatory states along that trajectory to govern the outcome of the stimulus identification process. We implemented a gradual reduction in the mean GC spiking threshold over the course of five sniffs of a corrupted odor signal, reflecting a concomitant increase in neuromodulator efficacy, and used the greatest of the five similarity values measured in the last gamma cycle within each sniff to classify the test odorant. Importantly, under very high noise conditions, each of the five "neuromodulatory" states performed best for some of the test odors and noise instantiations, indicating that a trajectory across a range of neuromodulatory states could yield superior classification performance compared to any single state. Indeed, this strategy yielded a substantial improvement in classification performance at very high levels of impulse noise, approximately doubling classification performance at P = 0.8 (Figure 4e).

In the biological system, olfactory bulb activity patterns resembling those evoked by a specific odor can be evoked by contextual priming that is predictive of the arrival of that odor [48]. We implemented this as a priming effect exerted by ascending piriform cortical neurons that synaptically excite GCs in olfactory bulb [49], the mapping between which can be learned dynamically [50]. Specifically, we presented the network with odor samples at an extreme level of destructive interference (P = 0.9) that largely precluded correct classification under default conditions (Figure 4e). When fractions of the population of GCs normally activated by the presented odor were primed by lowering their spike thresholds, classification performance improved dramatically, to a degree corresponding to the fraction of primed GCs (Figure 4f). That is, even a weak prior expectation of an incoming odor stimulus suffices to draw an extremely occluded odor signal out of the noise and into the attractor.

*Sample variance arising from plume dynamics*

In addition to occlusion by competing odorants, odor samples can vary based on the dynamics of their plumes (Figure 1c), which evolve over time. We therefore tested the algorithm's ability to recognize and classify samples of each odorant that were drawn from the wind tunnel at different points in time (Figure 5a-b). Specifically, in this paradigm, repeated samples of the same odorant differed from one another based on evolving odor plume dynamics, whereas samples of different odorants differed from one another both in plume dynamics and in the distribution of analyte sensitivities across the sensor array. Following one-shot training on all ten odors as described above, the spiking activity generated by odorant test samples was attracted over the five gamma cycles towards the corresponding learned representation. Notably, plume dynamics alone constituted a relatively minor source of variance compared to impulse noise (Figure 5c).

We then tested the network on samples incorporating both plume dynamics and impulse noise ($P = 0.4$). Following one-shot training on all ten odors, we sampled each odor across widely dispersed points in time, and contaminated each sniff with an independent instantiation of impulse noise (Figure 5d-e). Spiking activity was again attracted over the five gamma cycles of each sniff towards the correct learned representation (Figure 5f-h). Classification performance across levels of impulse noise from $P = 0.0$ to $P = 1.0$ (Figure 5i) indicated that the addition of plume-based variability moderately reduced network performance (compare to Figure 4e, *green curve*). Network performance was not affected by the introduction of noise correlations over time (Figure S1).

*Classification performance of the neuromorphic model*

To evaluate the performance of the EPL model, we compared its classification performance to the performance of multiple conventional signal processing techniques: a median filter (MF), a total variation filter (TVF; both commonly used as impulse noise reduction filters [51]), principal components analysis (PCA; a standard preprocessor used in machine olfaction [3]), and a seven-layer deep autoencoder (DAE; see Methods). Specifically, following training, we averaged the classification performance of each method across 100 different occluded presentations of each odor, with the occlusion level for each sample randomly and uniformly selected from the range $P = [0.2, 0.8]$, for a total of 1000 test samples. Incorrect classifications and failures to classify both were scored as failures.

The neuromorphic EPL substantially outperformed MF, TVF, and PCA. To model "one-sample" learning on the DAE for comparison with one-shot learning on the EPL network, we trained a DAE with one sample from each of the ten odorants over 1000 training epochs per odorant, with the odorants intercalated in presentation. The EPL network substantially outperformed the DAE under these conditions, in which the training set contained no information about the distribution of error that would arise during testing owing to impulse noise (Figure 6a). To improve DAE performance, we then trained it with 500 to 7000 samples of each of the ten odorants, with each sample independently occluded by impulse noise randomly and uniformly selected from the range $P = [0.2, 0.8]$. Under this training regimen, the deep network required 3000 samples per odorant, including the attendant information regarding the distribution of testing variance, to achieve the classification performance that the EPL model achieved with 1 sample per odorant. With further training, DAE performance exceeded that of the EPL network (Figure 6b). We then tested the online learning capacities of the two networks, in which the presentations of different odorants during training were sequential rather than uniformly interspersed. After training both networks to recognize toluene using the methods of Figure 6b, both the EPL and the DAE exhibited high classification performance. However, after subsequent training to recognize acetone, the DAE lost its ability to recognize toluene, whereas the EPL network recognized both odors with high fidelity (Figure 6c-d). Susceptibility to catastrophic forgetting is a well-established limitation of deep networks, though some customized networks recently have shown improvements in their online (continual) learning capabilities that reflect some of the strategies of the EPL network, such as the selective reduction of plasticity in well-trained network elements [42].

These results indicate that the EPL network ultimately serves a different purpose than techniques that require intensive training with explicit models of expected variance in order to achieve optimal performance. The EPL network is competitive with these algorithms overall, but excels at rapid, online learning with the capacity to generalize beyond experience in novel environments with unpredictable sources of variance. In contrast, the DAE evaluated here performs best when it is trained to convergence on data drawn from the distribution of expected variance; under these conditions, its performance exceeds that of the present EPL network. EPL network instantiations are thereby likely to be favored in embedded systems intended for deployment in the wild, where rapid training, energy-efficiency, robustness to unpredictable variance, and the ability to update training with new exemplars are at a premium.

**Discussion**

The EPL algorithm, while derived directly from computational features of the mammalian olfactory system, essentially comprises a spike timing-based variant of a Hopfield autoassociative network [52], exhibiting autoassociative attractor dynamics over sequential gamma-breadth packets of spiking activity. Since their conception, Hopfield networks and their variants have been applied to a range of computational problems, including sparse coding [53], combinatorial optimization [54], path integration [55], and oculomotor control [56]. Because these studies typically model neural activity as continuous-valued functions (approximating a spike rate), they have not overlapped significantly with contemporary research investigating spike-timing-based mechanisms of neural coding and computation [14, 17, 57-61] – mechanisms that are leveraged in contemporary neuromorphic systems to achieve massive parallelism and unprecedented energy efficiency [1, 62]. The EPL algorithm combines insights from these two bodies of work, instantiating autoassociative attractor dynamics within a spike timing framework. By doing so, it proposes novel functional roles for spike timing-dependent synaptic plasticity, packet-based neural communications, active neuromodulation, and adult neurogenesis, all instantiated within a scalable and energy-efficient neuromorphic platform (Figure 6f-g).

Contemporary artificial olfaction research often emphasizes the development of sensors and sensor arrays [63]. Associated work on the processing of electronic nose sensor data incorporates both established machine learning algorithms and novel analytical approaches [3, 4, 64], as well as optimizations for sensory sampling itself [65, 66]. The biological olfactory system has both inspired modifications of traditional analytical methods [4, 67] and guided biomimetic approaches to signal identification in both chemosensory and non-chemosensory datasets [10, 39, 68-73]. In comparison to these diverse approaches, the distinguishing features of the present report are the rapid learning of the EPL network, its spike timing-based attractor dynamics, its performance on identifying strongly occluded signals, and its field-deployable Loihi implementation.

**Summary**

Neuromorphic computing shows great promise, but presently suffers from a paucity of useful algorithms. Seeking inspiration from the circuit-level organization of biological neural systems, with their radically different computational strategies, provides a key opportunity to develop algorithmic approaches that might not otherwise be considered. We demonstrate that a simplified network model, based on the architecture and dynamics of the mammalian olfactory bulb [14] and instantiated in the Loihi neuromorphic system [1], can support rapid online learning and powerful signal restoration of odor inputs that are strongly occluded by contaminants. These results evince powerful computational features of the early olfactory network that, together with mechanistic models and experimental data, present a coherent general framework for understanding mammalian olfaction as well as improving the performance of artificial chemosensory systems. Moreover, this framework is equally applicable to other steady-state signal identification problems in which higher-dimensional patterns without meaningful lower-dimensional internal structure are embedded in highly interfering backgrounds.

**Acknowledgments**

*Funding*

Supported by NIDCD R01 grants DC014701 and DC014367 to TAC.

*Author Contributions*

TAC originally conceived the neural algorithm; NI and TAC developed the algorithm to practice. NI instantiated the algorithm and enhanced the learning rules, and subsequently ported it to Loihi. TAC and NI wrote the manuscript.

*Data and Materials Availability*

The Vergara et al. gas sensor dataset [4] is freely available from the UCI Machine Learning database (http://archive.ics.uci.edu/ml/datasets/gas+sensor+arrays+in+open+sampling+settings).

**Methods**

*Dataset and odorant sampling*

Sensory input to the model was generated from the "Gas sensor arrays in open sampling settings" dataset published by Vergara and colleagues [4] and available from the UCI Machine Learning Repository (https://bit.ly/2nteY78). The dataset comprises the responses of 72 metal-oxide based chemical sensors distributed across a wind tunnel. There are six different sensor mounting locations in the tunnel, three different settings of the tunnel's wind speed and three different settings of the sensor array's heater voltage. In our present study, we consider the recordings made at sensor location "L4" (near the mid-point of the tunnel), with the wind speed set to 0.21 m s$^{-1}$ and the heater voltage set to 500 V. The tunnel itself was 1.2 m wide x 0.4 m tall x 2.5 m long, with the sensors deployed in nine modules, each with eight different sensors, distributed across the full 1.2 m width of the tunnel at a location 1.18 m from the inlet (Figure 1c). The nine modules were identical to one another. To maintain the generality of the algorithm rather than optimize it for this particular dataset, we here sampled the 72 sensors naively, without in any way cross-referencing inputs from the nominally identical sensors replicated across the nine modules, or attempting to mitigate the plume-based variance across these sensors. The turbulent plume shown in Figure 1c is illustrative only; distribution maps of local concentrations in the plume, along with full details of the wind tunnel configuration, are provided in the publication first presenting the dataset [4].

Ten different odorants were delivered in the gas phase to the sensor array: acetone, acetaldehyde, ammonia, butanol, ethylene, methane, methanol, carbon monoxide, benzene, and toluene [4]. For every tunnel configuration, each of these odorants was presented 10-20 times, and each presentation lasted for 180 seconds. In our present study, we consider one of these 180-second plumes (chosen at random) for each odorant.

We discretized each sensor's range of possible responses into 16 levels of activation, corresponding to 16 time bins of the permissive epoch of each gamma cycle (see *Intrinsic gamma and theta dynamics section*). The discretized sensor values were composed into a 72-dimensional sensor activity vector, which then was sparsened by setting the smallest 50% of the values to zero. Accordingly, each odorant sample ("sniff") presented to the EPL network comprised a discrete 72-element sensor vector drawn from a single point in time and presented as steady state. The training set underpinning one-shot learning was based on single-timepoint samples drawn from the 90 second timepoint in each of the 180 second long odorant presentations. Test sets for the impulse-noise-only studies (Figures 3-4) comprised these same timepoints, each altered by 100 different instantiations of impulse noise. For the plume-variance studies (Figure 5), test samples for each odorant were drawn from different time points in the corresponding plume (specifically, across the range 30-180 seconds after odorant presentation, at 5 second intervals) and were presented to the network both with and without added impulse noise.

The OB EPL model therefore was instantiated with 72 columns, such that each column received afferent excitation proportional to the activation level of one sensor. Because we here present the network in its simplest form, we treated the 72 columns as independent inputs, without crafting the algorithm to combine the responses of duplicate sensor types, to weight the centrally located sensors more strongly, or to perform any other dataset-specific modifications that might improve performance. Each model OB column comprised one principal neuron (MC) and initially five GC interneurons that were presynaptic to that MC (for a total of 360 GCs across all columns), though the number of GCs per column rose as high as 50 in the most highly trained models described herein (see *Adult neurogenesis* section). MCs projected axons globally across all columns and formed excitatory synapses onto GCs with a uniform probability of 0.2 (20%). Each GC, in turn, synaptically inhibited the MC within its column with a probability of unity (100%). GCs did not inhibit MCs from other columns, though this constraint can be relaxed without affecting overall network function. To reflect the mapping of the algorithm to the physical layout of the Loihi chip, we consider an MC and its co-columnar GCs to be spatially local to one another. However, there is no computational basis for the physical locations of neurons in the model; an OB column is simply "an MC plus those inputs that can affect its activity".

*Intrinsic gamma and theta dynamics*

In the biological system, the profile of spike times across MCs is proposed to reflect a phase precedence code with respect to the emergent gamma-band field potential oscillations generated in the olfactory system. Spike timing-based coding metrics are known to offer considerable speed and efficiency advantages [57-61]; however, they require computational infrastructure in the brain to realize these benefits. Fast oscillations in the local field potential are indicative of broad activity coherence across a synaptically coordinated ensemble of neurons, and thereby serve as temporal reference frames within which spike times in these neurons can be regulated and decoded. Accordingly, these reference frames are essential components of the biological system's computational capacities.

In the OB, gamma oscillations emerge from interactions of the subthreshold oscillations of MCs with the network dynamics of the EPL (PRING dynamics [14]). For present purposes, the importance of these oscillations was twofold: (1) MC spike phases with respect to the gamma-band oscillations serve as the model's most informative output, and (2) by considering each oscillation as embedding a distinct, interpretable representation, repeated oscillations enable the network to iteratively approach a learned state based on stationary sensory input. Notably, in vivo, piriform cortical pyramidal neurons are selectively activated by convergent, synchronous MC spikes [74], and established neural learning rules are in principle capable of reading such a coincidence-based metric [75]. Because MC spike times can be altered on the gamma timescale by synaptic inhibition from GCs, and their spike times in turn alter the responsivity of GCs, these lateral inhibitory interactions can iteratively modify the information exported from the OB. In the neuromorphic EPL, each MC periodically switched between two states to establish the basic gamma oscillatory cycle. These two states were an active state in which the MC integrated sensory input and generated spikes (*permissive epoch*) and an inactive state in which the excitation level of the MC was held at zero, preventing sensory integration and spike generation (*inhibitory epoch*; Figure 1d). The effects of the plastic lateral inhibitory weights from GCs were applied on top of this temporal framework (see *Inhibitory synaptic plasticity* section). The correspondence with real time is arbitrary and hence is measured in timesteps (ts) directly; that said, as Loihi operates at about 100 kHz, each timestep corresponds to about 10 us. In the present implementation, the permissive epoch comprised 16 ts and the inhibitory epoch 24 ts, for a total of 40 ts per gamma cycle. Notably, the duration of the permissive epoch directly corresponds to the number of discrete levels of sensory input that can be encoded by our spike timing-based metric; it can be expanded arbitrarily at the cost of greater time and energy expenditures.

A second, slower, sampling cycle was used to regulate odor sampling. This cycle is analogous to theta-band oscillations in the OB, which are driven primarily by respiratory sampling (sniffing) behaviors but also by coupling with other brain structures during certain behavioral epochs. Each sampling cycle ("sniff") consisted of a single sample and steady-state presentation of sensory input across five gamma cycles of network activity. The number of gamma cycles per sampling cycle can be arbitrarily determined in order to regulate how much sequential, iterative processing is applied to each sensory sample, but was held at five for all experiments herein.

Importantly, these differences between the slower sampling timescale and the faster processing timescale can be leveraged to implement "continuous" online sampling, in which each sample can be processed using multiple computational iterations prior to digitizing the next sample. In the present implementation, for example, the Vergara et al. dataset sampled odorants at 100 Hz – one sample every 10 ms. On Loihi, operating at 100 kHz, the 200 timesteps (5 gamma cycles) used for the processing of a single sniff require a total of around 2 ms. As this is five times faster than the sampling rate of the sensors, there would be no update to sensor state during the time required for five cycles of processing. Examples of this algorithm operating in "continuous" mode are presented in supplemental Figure S1.

*Mitral cells*

Each MC was modeled by two compartments – an apical dendrite (AD) compartment that integrated sensor input and generated "spike initiation" events when an activation threshold was crossed, and a soma compartment that was excited by spike initiation events in the AD compartment and synaptically inhibited by spikes evoked in cocolumnar GCs. The soma compartment propagated the AD-initiated spike as an MC action potential after release from GC inhibition. Accordingly, stronger sensory inputs initiated earlier (phase-leading) spikes in MCs, but the propagation of these spikes could be delayed by inhibition arising from presynaptic GCs. Distinguishing between these two MC compartments facilitated management of the two input sources and their different coding metrics, and reflected the biophysical segregation between the mass-action excitation of MC dendritic arbors and the intrinsic regulation of MC spike timing governed by the gamma-band oscillatory dynamics of the OB external plexiform layer [14].

Sensor activation levels were delivered to the AD compartment of the corresponding column, which integrated the input during each permissive epoch of gamma. If and when the integrated excitation exceeded threshold, a spike initiation event was generated and communicated to the soma compartment. Stronger inputs resulted in more rapid integration and correspondingly earlier event times. After generating an event, the AD was not permitted to initiate another for the duration of that permissive epoch.

A spike initiation event in the AD generated a unit level of excitation (+1) in the soma compartment for the remainder of the permissive epoch. This excitation state caused the MC soma to propagate the spike as soon as it was sufficiently free of lateral inhibition received from its presynaptic GCs. Accordingly, the main effect of GC synaptic inhibition was to modulate MC spike times with respect to the gamma cycle. The resulting MC spikes were delivered to the classifier as network output, and also were delivered to its postsynaptic GCs.

During the first gamma cycle following odor presentation, when GC inhibition was not yet active, the soma immediately propagated the MC spike initiated in the AD. After propagating a spike, the soma was not permitted to spike again for the duration of the permissive epoch. At the end of the permissive epoch, both the AD and soma compartments were reset to zero for the duration of the inhibitory epoch.

*Granule cells*

GCs were modeled as single-compartment neurons,

$$V = \sum_k w_k s_k \qquad (1)$$

in which V indicates the excitation level of the GC, $w_k$ represents the excitatory synaptic weight from a presynaptic MC soma $k$, and $k$ was summed over all presynaptic MCs. The boolean term $s_k$ denotes a spike at the $k$-th presynaptic MC soma; $s_k$ equals 0 at all times except for the $d$-th timestep following a spike in the k-th MC soma, when it was set to 1. Accordingly, $d$ denotes a delay in the receipt of synaptic excitation by a GC following an MC spike. This delay $d$ was randomly determined, synapse-specific, and stable (i.e., not plastic); it reflects heterogeneities in spike propagation delays in the biological system and served to delay GC excitation such that GC spikes were evoked within the inhibitory epoch of gamma.

A spike in an MC soma $k$ that was presynaptic to a given GC excited that GC in proportion to its synaptic weight $w_k$. Once GC excitation rose above a threshold $\theta_{GC}$, the GC generated a spike and reset its excitation level to zero. Following a spike, the GC was not permitted to spike again for 20 timesteps, ensuring that only one spike could be initiated in a given GC per gamma cycle. In general, convergent excitation from multiple MCs was required for GC spike initiation.

*Excitatory synaptic plasticity*

The weights of MC-to-GC synapses were initialized to a value of $w_e$. Following an asymmetric, additive spike timing-dependent plasticity rule, these synaptic weights were modified during training following a spike in the postsynaptic GC. Specifically, synapses in which the presynaptic MC spike preceded the postsynaptic GC spike by 1 timestep were potentiated by a constant value of $\delta_p$ whereas all other synapses were depressed by a constant value of $\delta_d$. In the present study, we set $\delta_p$ to $0.05w_e$ and $\delta_d$ to $0.2w_e$. GC spiking thresholds were set to $6w_e$.

The overall effect of this rule was to develop sparse and selective higher-order receptive fields for each GC, a process termed *differentiation*. Specifically, repeated coincidences of the same MC spikes resulted in repeated potentiation of the corresponding synapses, whereas synapses of other MCs underwent repeated depression. Individual excitatory synaptic weights were capped at a value of $1.25w_e$, ensuring that the spiking of differentiated GCs remained sensitive to coincident activity in a particular ensemble of MCs, the number of which constituted the order of the GC receptive field. By this process, odor learning transformed the relatively broad initial receptive field of a GC into a highly selective one of order M. These higher-order receptive fields reflected correlations between components of individual sensor vectors – i.e., the higher-order signatures of learned odors. Differentiated GCs thereby developed selectivity for particular odor signatures and became unresponsive to other sensory input combinations. While in principle this GC output can be used directly for classification purposes [39, 40], the present algorithm instead deploys it to denoise the spike timing-based MC representation. Because there are many fewer MCs than GCs, there is a corresponding reduction in bandwidth and energy consumption by using MCs to communicate the representation for classification or further processing.

*Adult neurogenesis*

The process of GC differentiation permanently depleted the pool of interneurons available for recruitment into new odor representations. To avoid a decline in performance as the numbers of odors learned by the network increased, we periodically added new, undifferentiated GC interneurons to the network on a timescale slower than that of the synaptic plasticity rules – a process directly analogous to adult neurogenesis in olfactory bulb [26, 27, 33, 34, 43]. Specifically, the network was initialized with five GCs per column, as described above. After the learning of each new odor, an additional set of five undifferentiated GCs was configured in every column. As with the initial network elements, every MC in the network formed excitatory synapses onto new GCs with a probability of 0.2 (20%), and the new GCs all formed inhibitory synapses onto their cocolumnar MCs with initial inhibitory weights of zero.

*Inhibitory synaptic plasticity*

In the neuromorphic model, inhibitory synapses from presynaptic GCs onto their cocolumnar MC somata exhibited three functional states. The default state of the synapse was an inactive state I, which exerted no effect on the MC (i.e., equal to 0). When a spike was evoked in the GC, the synapse transitioned into an inhibitory blocking state B; this state was maintained for a period of time $\Delta_B$ that was determined by learning. While in this state, the synapse maintained a unit level of inhibition (equal to –1) in the postsynaptic MC soma that inhibited somatic spike propagation. The blocking period $\Delta_B$ therefore governed MC spike latency, and corresponded functionally to the inhibitory synaptic weight. At the end of the blocking state, the synapse transitioned to a release state R for 1 timestep, during which it generated a unit level of excitation (equal to +1) in the postsynaptic MC soma. The synapse then resumed the inactive state. An MC soma propagated a spike when the sum of the excitation and inhibition generated by its apical dendrite and by the synapses of all of its presynaptic GCs was positive. After spiking once, the MC soma was not permitted to spike again for the duration of that gamma cycle.

All inhibitory synaptic weights in new GCs were initialized to $\Delta_B = 0$ ts. During training, additionally, the effects of inhibition on MC somata were suppressed. If an MC AD initiated a spike within the permissive epoch immediately following a cocolumnar GC spike (in the previous inhibitory epoch), the blocking period $\Delta_B$ imposed by that GC onto the soma of that MC was modified based on the learning rule

$$\delta_b = \eta(t_{AD} - t_R) \qquad (2)$$

where $\delta_b$ is the change in the blocking period $\Delta_B$ (inhibitory synaptic weight), $t_{AD}$ is the time of the MC spike initiation event in the AD, $t_R$ is the time at which the inhibitory synapse switched from the blocking state to the release state, and $\eta$ was the learning rate (set to 1.0 in the one-shot learning studies presented here). Consequently, the synaptic blocking period $\Delta_B$ was modified during training (rounding up fractions) until the release of inhibition from that synapse was aligned with the spike initiation event in the MC AD (Figure 2c). If the GC spike was not followed by an MC spike initiation event during the following permissive epoch, the inhibitory weight $\Delta_B$ of that synapse grew until that MC was inhibited for the entire gamma cycle. Inputs from multiple local GCs onto a common MC were applied and modified independently.

In total, this inhibitory synaptic plasticity rule enabled the EPL network to learn the timing relationships between GC spikes and cocolumnar MC spikes associated with a given odor stimulus, thereby training the inhibitory weight matrix to construct a fixed-point attractor around the odor representation being learned. This served to counteract the consequences of destructive interference in odor stimuli presented during testing. Importantly, this plasticity rule effectively learned the specific ratiometric patterns of activation levels among MCs that characterized particular odors; consequently, two odors that activated the same population of MCs, but at different relative levels, could be readily distinguished.

*Testing procedures*

After training, we tested the network's performance on recognizing learned odorants in the presence of destructive interference from unpredictable sources of olfactory occlusion (impulse noise), alone or in combination with variance arising from sampling plume dynamics at different timepoints. All testing was performed with network plasticity disabled.

The responses of primary olfactory receptors to a given odorant of interest can be radically altered by the concomitant presence of competing background odorants that strongly activate or block some of the same receptors as the odorant of interest, greatly disrupting the ratiometric activation pattern across receptors on which odor recognition depends. We modeled this occlusion as destructive impulse noise. Specifically, an occluded test sample was generated by choosing a fraction *P* of the 72 elements of a sensor activity vector and replacing them each with random values drawn uniformly from the sensors' operating range (integer values from 0 to 15). When multiple occluded test samples were generated to measure average performance, both the identities of the occluded elements and the random values to which they were set were redrawn from their respective distributions.

Odor plume dynamics comprise a second source of stimulus variance encountered under natural conditions. To test network performance across this variance, we drew test samples from different timepoints within the odor plumes. Specifically we drew 30 samples per plume at 5 second intervals between 30 seconds and 180 seconds within the 180 second datastreams. After one-shot training with a single sample, we tested network performance on the other samples, with and without the addition of impulse noise (Figure 5).

While the present study focuses on one-shot learning, the network can also be configured for few-shot learning, in which it gradually adapts to the underlying statistics of training samples. In this configuration, the network learns robust representations even when the training samples themselves are corrupted by impulse noise. We illustrate this effect in supplementary Figure S2.

*Sample classification*

The pattern of MC spikes in each successive gamma cycle was recorded as a set of spikes, with each spike defined by the identity of the active MC and the spike latency with respect to the onset of that permissive epoch. Accordingly, five successive sets of spikes were recorded for each sample "sniff". When an impulse noise-occluded sample was presented to the network, the similarities were computed between each of the five representations evoked by the unknown and each of the network's learned odor representations. In descriptive figures (but not for comparisons with other methods), the similarity between two representations was measured with the Jaccard index, defined as the number of spikes in the intersection of two representations, divided by the number of spikes in their union [44]. Specifically, the permissive epoch of a gamma cycle included 16 discrete timesteps in which MCs could spike; these 16 bins were used for Jaccard calculations. Test samples were classified as one of the network's known odorants if the similarity exceeded a threshold of 0.75 in the fifth (final) gamma cycle. If similarities to multiple learned odorants crossed the threshold, the odorant exhibiting the greatest similarity value across the five gamma cycles was picked as the classification result. If none of the similarity values crossed the threshold within five gamma cycles, the odorant was classified as unknown. This combination of nearest-neighbor classification and thresholding enabled the network to present "none of the above" as a legitimate outcome. Summary figures each consist of averages across 100 independent instantiations of impulse noise, and/or averages across 30 different test samples drawn from different timepoints in the datastream (without or with added impulse noise), for each odor in the training set.

*Benchmarks*

We first compared the classification performance of the EPL network to three conventional signal processing techniques: a median filter (MF), a total variation filter (TVF), and principal component analysis (PCA; Figure 6a) [3, 51]. The MF and TVF are filters commonly used in signal processing for reducing impulse noise, while PCA is a standard preprocessor used in machine olfaction applications [3]. The MF used a window size of 5, and was implemented with the Python signal processing library *scipy.signal*. The TVF used a regularization parameter equal to 0.5, and was implemented using the Python image processing library *scikit-image*. PCA was implemented using the Python machine learning library *scikit-learn*; data were projected onto the top five components.

Corrupted input signals also can be denoised by training an autoencoder, a modern rendition of autoassociative networks [76, 77]. We therefore compared the performance of the EPL network to a seven-layer deep autoencoder constructed using the Python deep learning library *Keras*. The seven layers consisted of an input layer of 72 units, followed by five hidden layers of 720 units each and an output layer of 72 units. This resulted in a network of 3744 units, identical to the number in the EPL model when trained with ten odors. The network was fully connected between layers, and the activity of each unit in the hidden layers was L1 regularized. The network was trained with iterative gradient descent until convergence using the Adadelta optimizer with a mean absolute error loss function. Its training set consisted of 7000 examples per odorant class. For the same training set, the performance of this seven-layer autoencoder exceeded that of shallower networks (6-, 5-, 4-, and 3-layer networks were tested).

For direct comparison, the outputs of all of these methods, including that of the EPL network, were presented to the same nearest-neighbor classifier for sample classification according to a Manhattan distance metric. Specifically, for each of the techniques, the output was read as a 72-dimensional vector and normalized such that their elements summed up to a value of unity. (In the case of the EPL network, the spiking output in each gamma cycle was read out as a 72-dimensional rank-order vector and normalized so that the elements summed to unity). The similarity between any two such vectors was measured as $(1/(1+d))$ where $d$ is the Manhattan distance between the two vectors. Classification performance was measured by computing this similarity between the output of training data samples and those of test data samples. A test data sample was classified according to the identity of the training data sample to which it was most similar, provided that this similarity value exceeded a threshold of 0.75 (thresholding enabled the inclusion of a "none of the above" outcome).

We trained the deep autoencoder (DAE) in three different ways for fair comparison with EPL network performance. First, the DAE was trained using the same ten non-occluded odor samples that were used to train the EPL model. These ten samples underwent 1000 training epochs to ensure training convergence. This method assesses DAE performance on "one-sample" learning, for comparison with the one-sample/one-trial learning of the EPL network (Figure 6a). Second, we trained the DAE on multiple impulse noise-occluded samples, so as to maximize its performance. Specifically, we trained the DAE on 500 to 7000 training samples, where each sample comprised an independently occluded instance of each of the ten odorants. Each training set was presented for 25 training epochs to ensure convergence. The occlusion levels for each training sample were drawn from the same distribution as the test samples, being randomly and uniformly selected from the range P = [0.2, 0.8]. With this procedure, we show that the DAE requires 3000 training samples per odorant to achieve the classification performance that the EPL model achieved with 1 training sample per odorant (Figure 6c); i.e., the EPL model is 3000 times more data efficient than the DAE. Third, we trained the DAE and EPL models first on one odorant (toluene) and then, subsequently, on a second odorant (acetone) in order to compare the models' sequential online learning capabilities. After training on toluene, the DAE classified test presentations of toluene with high fidelity (Figure 6c; *left panel*). However, over the course of acetone training, the similarity between test samples of toluene and the learned representation of toluene progressively declined (Figure 6d), to the point that the DAE network became unable to correctly classify toluene (Figure 6c, *right panel*; see [42] for strategies to improve the sequential learning performance of deep networks). In contrast, training the EPL network with acetone exhibited no interference with the preexisting toluene representation (Figure 6d, *inset*). The similarity between test samples of toluene and the learned representation of toluene was not affected as the EPL learned all of the ten odorants in sequence (Figure 6e).

*Implementation on the Loihi neuromorphic system*

Neuromorphic systems are custom integrated circuits that model biological neural computations, typically with orders of magnitude greater speed and energy efficiency than general-purpose computers. These systems enable the deployment of neural algorithms in edge devices, such as chemosensory signal analyzers, in which real-time operation, low power consumption, environmental robustness, and compact size are critical operational metrics. Loihi, a neuromorphic processor developed for research at Intel Labs, advances the state of the art in neuromorphic systems with innovations in architecture and circuit design, and a feature set that supports a wide variety of neural computations [1]. Below we provide an overview of the Loihi system and our network implementation thereon.

Loihi is fabricated in Intel's 14-nm FinFET process and realizes a total of 2.07 billion transistors over a manycore mesh. Each Loihi chip contains a total of 128 neuromorphic cores, along with three embedded Lakemont x86 processors and external communication interfaces that enable the neuromorphic mesh to be extended across many interlinked Loihi chips (Figure 1b). Each neuromorphic core comprises leaky-integrate-and-fire compute units that integrate filtered spike trains from a configurable set of presynaptic units and generate spikes when a threshold level of excitation is crossed. Postsynaptic spikes then are communicated to a configurable set of target units anywhere within the mesh. A variety of features can be configured in a core, including multicompartment interactions, spike timing-dependent learning rules, axonal conduction delays, and neuromodulatory effects. All signals in the system are digital, and networks operate as discrete-time dynamical systems.

We configured each column of our model within one neuromorphic core, thereby using a total of 72 cores on a single chip. Cocolumnar synaptic interactions took place within a core, whereas the global projections of MC somatic spikes were routed via the intercore routing mesh. The configured network utilized 12.5% of the available neural resources per core and 6% of the available synaptic memory.

Completing one inference cycle (sniff; 5 gamma cycles; 200 timesteps) of the 72-core network required 2.75 ms and consumed 0.43 mJ, of which 0.12 mJ is dynamic energy. Critically, the time required to solution was not significantly affected by the scale of the problem (Figure 6f), owing to the Loihi architecture's fine-grained parallelism. This scalability highlights a key advantage of neuromorphic hardware for application to computational neuroscience and machine olfaction. Energy consumption also scaled only modestly as network size increased (Figure 6g), owing to the colocalization of memory and compute and the use of sparse (spiking) communication, which minimize the movement of data. Using multichip Loihi systems, we envision scaling up the present implementation to hundreds of columns and hundreds of thousands of interneurons, as well as to integrate circuit models of the glomerular layer [78] and the piriform cortex with the current EPL network of olfactory bulb.

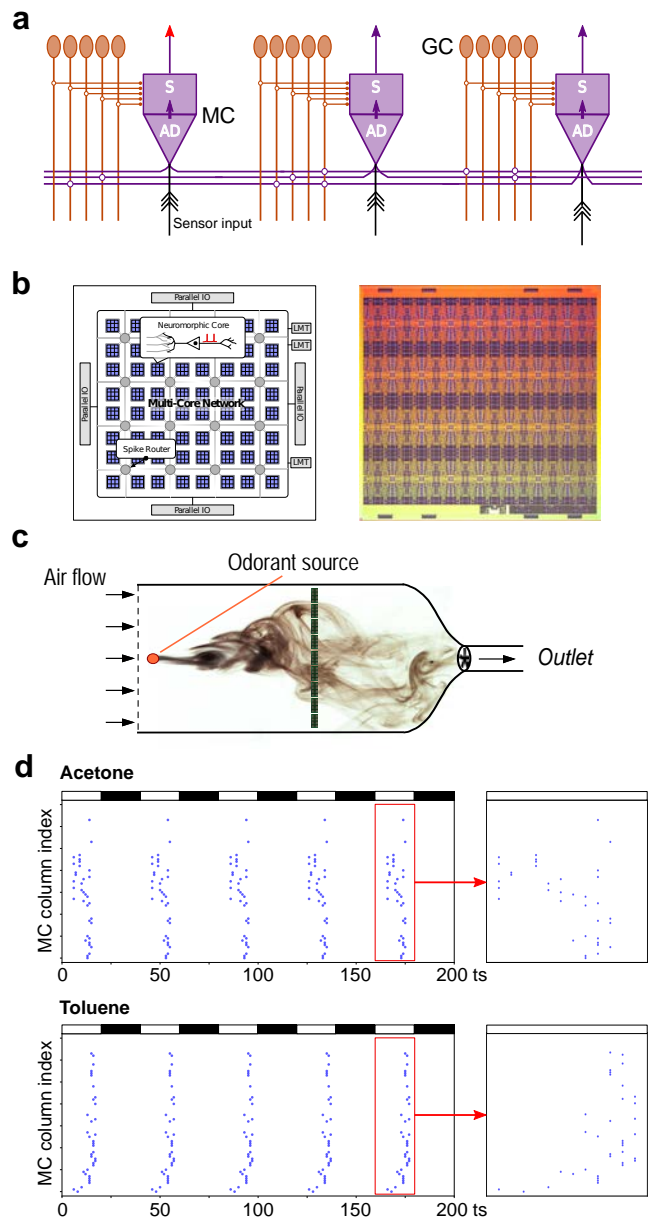

**Figure 1.** Model structure and signal encoding. *a,* Architecture of the neuromorphic model. Sensor input is delivered to the apical dendrite (AD) of each mitral cell (MC), which in turn excites its corresponding soma (S). The resulting MC activation is propagated out via its lateral dendrites (*purple*) to synaptically excite the dendrites of granule cells (GC, *orange*). The distribution of excitatory connections (*open circles*) is sparse and independent of spatial proximity. In contrast, GC spiking activity is delivered as inhibition onto its local, cocolumnar MC. *b,* Architecture of the Intel Loihi neuromorphic chip [1]. *Left panel.* Neuromorphic cores (blue squares) operate in parallel and communicate through a mesh of spike routers (grey circles). Also depicted are the three embedded x86 Lakemont cores (LMT) and the input/output interfaces (IO). *Right panel.* The Loihi silicon, fabricated in Intel's 14nm FinFET process and consisting of 2.07 billion transistors over the manycore mesh. *c,* Illustration of odorant delivery to a 72-element chemosensor array within a wind tunnel [4]. Nine banks of eight sensors each were deployed across the full 1.2 m breadth of the tunnel. *d,* Presentation of acetone (*top*) or toluene (*bottom*) to the chemosensor array resulted in characteristic patterns of spiking activity across the 72 MCs (*ordinate*). Stronger sensor activation led to correspondingly earlier MC spikes within each gamma cycle. In the absence of noise, the response was odor-specific but stationary across five sequential gamma cycles. Adjacent indices refer to sensors that are in adjacent locations across the wind tunnel, irrespective of their type. Inhibitory epochs (20 timesteps duration) are denoted by black bars; permissive epochs (20 timesteps) are denoted by white bars. The fifth gamma cycle is expanded in time (*rightmost panels*) to illustrate the distribution of MC spike times. *ts,* timesteps.

**Figure 2**

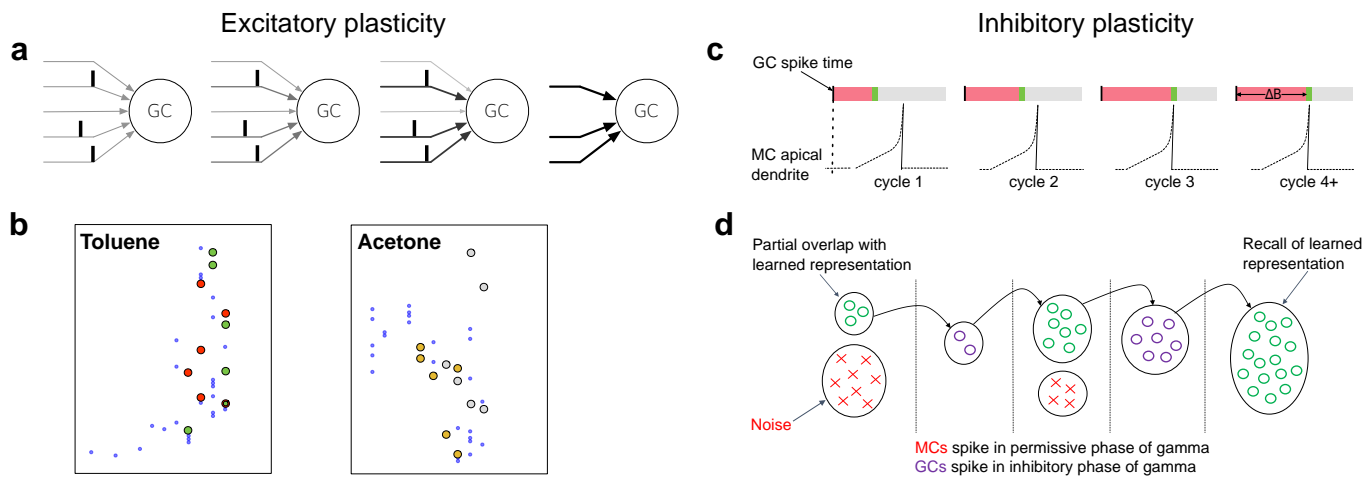

**Figure 2**. Plasticity rules. *a,* During training, repeated coincident MC spikes converging onto a given GC activated that GC, and developed strong excitatory synaptic weights thereon, whereas other inputs to that GC were weakened and ultimately eliminated. *b,* Excitatory plasticity rendered GCs selective to higher-order features of odor representations. After training on toluene (*left panel*) or acetone (*right panel*), a number of GCs became responsive to specific combinations of activated MCs. Spike times highlighted with *green* spots denote those MC spikes that activated a specific GC in a network trained on toluene. *Red* spots denote a second such GC, responsive to toluene via a different set of activated MCs. *Yellow* and *grey* spots denote analogous MC spike populations that activate two GCs responsive to acetone in the same network. *c,* Illustration of the inhibitory plasticity rule. During training, the weight (duration) of spike-mediated GC inhibition onto its cocolumnar MC (*red bar*) increased until the release of this inhibition (*green*) coincided with spike initiation in the MC apical dendrite. The learned inhibitory weight corresponded to a blocking period ΔB during which spike propagation in the MC soma was suppressed. *d,* Illustration of the iterative denoising of an occluded test sample. Partially-correct representations in MCs evoke responses in some of the correct GCs, which deliver inhibition that draws MC ensemble activity iteratively closer to the learned representation. Three permissive epochs interspersed with two inhibitory epochs are depicted.

**Figure 3**

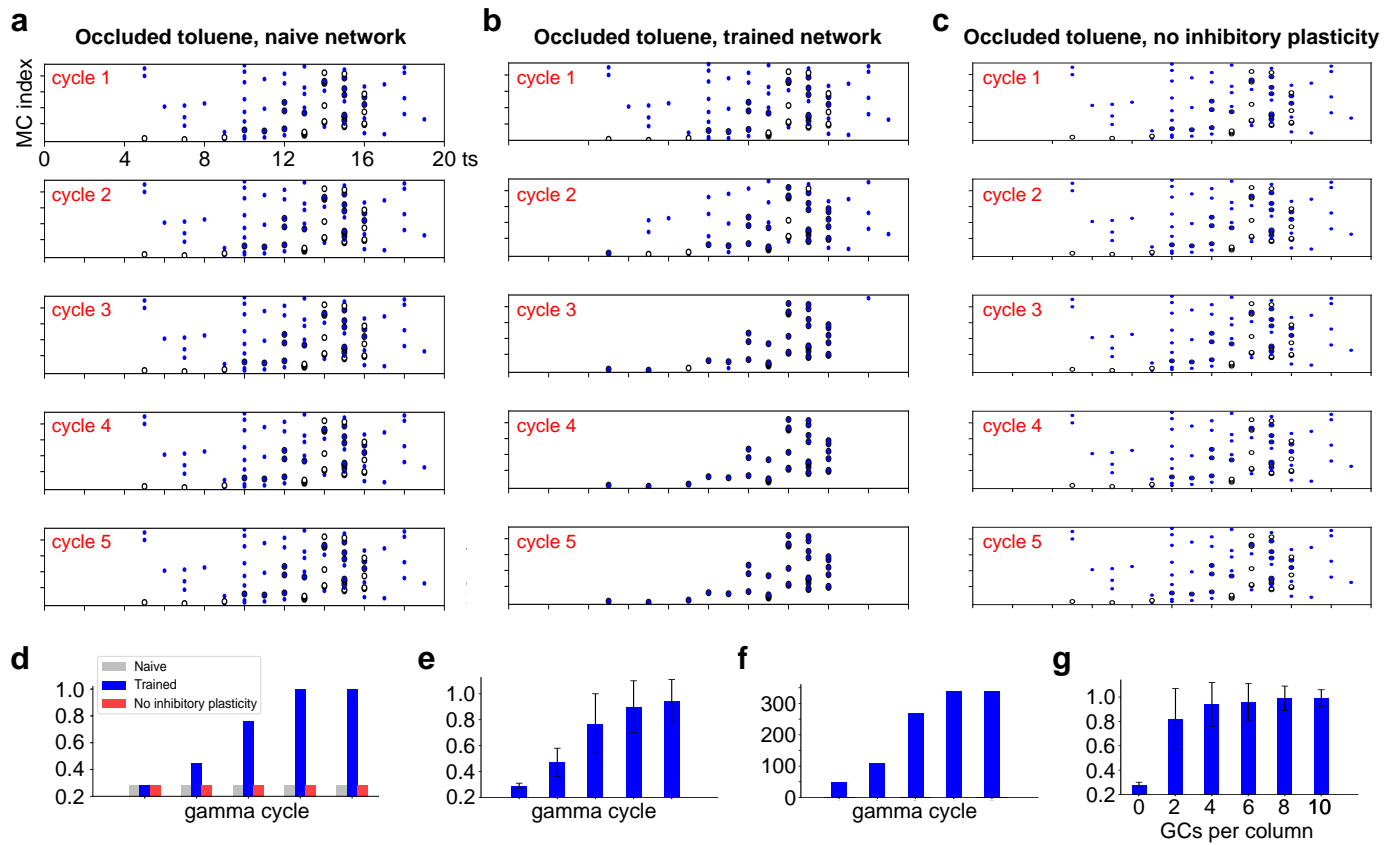

**Figure 3**. Odorant-evoked MC activity patterns are attracted to learned representations. *a,* Presentation of an occluded instance of toluene to an untrained network. Blue dot rasters denote spike times evoked by occluded toluene (impulse noise P = 0.6). The 72 rows are ordered according to sensor locations across the wind tunnel. The untrained network does not update the response to occluded toluene over the five gamma cycles depicted. For comparison, open circle rasters denote the spike times evoked by non-occluded toluene. *ts*, timesteps. *b,* Presentation of the same occluded instance of toluene to a plastic network trained on (non-occluded) toluene. The activity profile evoked by the occluded sample was attracted to the learned toluene representation over successive gamma cycles. *c,* Presentation of the same occluded instance of toluene to a network trained on non-occluded toluene with excitatory, but not inhibitory, plasticity enabled. The omission of inhibitory plasticity rendered the network unable to denoise MC representations during testing. *d,* The Jaccard similarity [44] between the response to occluded toluene and the learned representation of toluene systematically increased over five gamma cycles in the trained network (panel b), but not in the untrained network (panel a) or the network with inhibitory plasticity disabled (panel c). *e,* The Jaccard similarity increased reliably over five gamma cycles when averaged over 100 independently generated instances of occluded toluene (impulse noise P = 0.6). Error bars denote standard deviation. *f,* During learning, the number of GCs tuned to toluene increased over the five successive gamma cycles of training. *g,* Mean Jaccard similarity in the fifth gamma cycle as a function of the number of undifferentiated GCs per column. Mean similarity is averaged across 100 occluded instances of toluene (impulse noise P = 0.6); error bars denote standard deviation. Five GCs per column were utilized for all other simulations described herein.

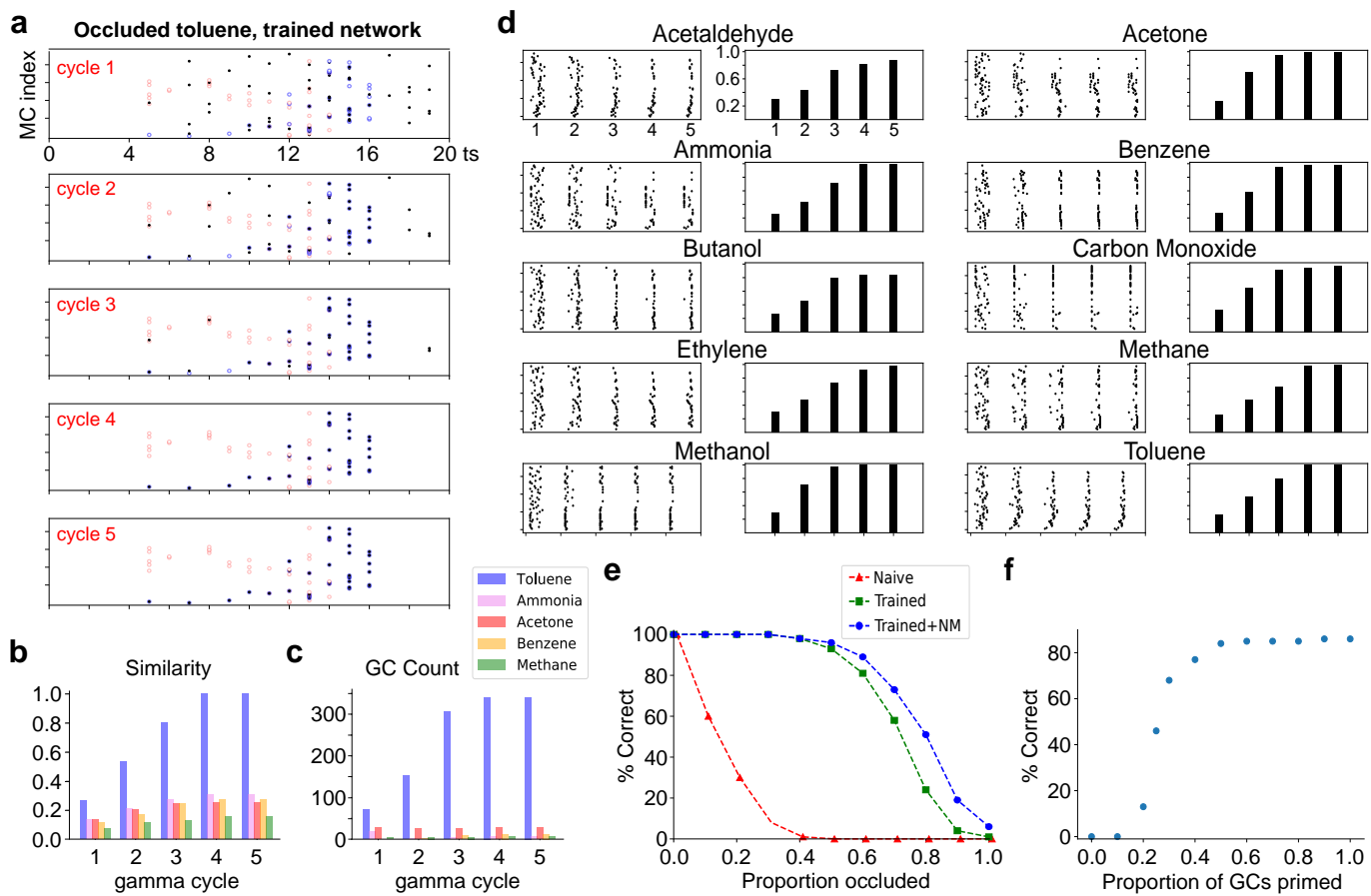

**Figure 4**. Multi-odor learning. *a,* Spike raster plot depicting attractor dynamics after training the network on all ten odorants. The representation generated by a sample of occluded toluene ($P = 0.6$; *black dots*) was progressively drawn towards the learned representation of toluene (*open blue circles*) and away from the learned representations of acetone (*open red circles*) and the other eight odorants (not shown). *ts*, timesteps. *b,* The Jaccard similarity to toluene that was evoked by the occluded-toluene stimulus increased over five successive gamma cycles until the stimulus was classified as toluene (similarity > 0.8). For clarity, only five odorants are depicted. *c,* The number of toluene-tuned GCs activated by the occluded-toluene stimulus progressively increased over five gamma cycles as the MC spiking activity pattern was attracted towards the learned toluene representation. GCs tuned to the other nine odorants were negligibly recruited by the evolving stimulus representation. *d,* Network activity evoked by presentation of occluded instances of each of the ten learned odors following one-shot learning. *Left panels*, spike raster plots over five gamma cycles (200 timesteps); *ordinate* denotes MC index. *Right panels*, Jaccard similarity between the activity pattern generated by each occluded odorant stimulus and the learned representation of the corresponding odorant. The same network can reliably recognize all ten odorants from substantially occluded examples ($P = 0.6$). *e,* Mean classification performance across all ten odorants under increasing levels of sensory occlusion (100 impulse noise instantiations per odorant per noise level). The abscissa denotes the level of impulse noise – i.e., the proportion of MC inputs for which the sensory activation level was replaced with a random value. *Red curve*, proportion of correct classifications by an untrained network. *Green curve*, proportion of correct classifications by a network trained on all ten odorants. *Blue curve*, proportion of correct classifications by a trained network with the aid of a neuromodulation-dependent dynamic state trajectory. *f,* Effects of GC priming on classification performance

under extreme occlusion. One hundred independently generated samples of occluded toluene with impulse noise P = 0.9 were presented to the fully-trained network. The putative effects of priming arising from piriform cortical projections onto bulbar GCs were modeled by lowering the spike thresholds of a fraction of toluene-tuned GCs. As the fraction of toluene-tuned GCs so activated was increased, classification performance increased from near zero to over 80% correct.

**Figure 5**

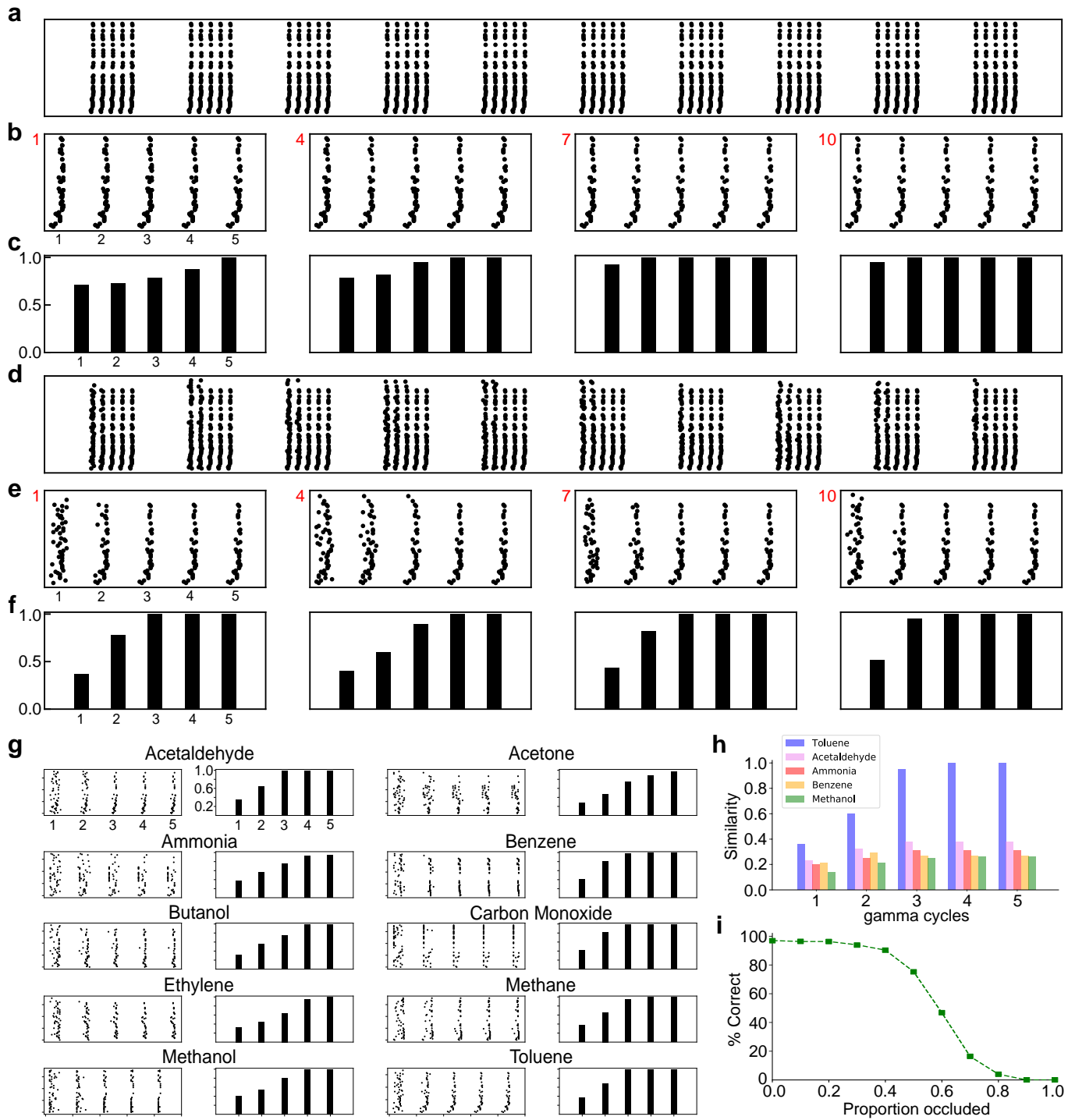

**Figure 5**. Odor learning with plume dynamics. *a*, Ten sniffs of toluene drawn from randomly-selected timepoints within the dataset illustrate sampling variance arising from plume dynamics. *Ordinate* denotes MC index, ordered according to sensor locations across the wind tunnel. *b*, Higher-resolution depictions of sniffs #1, 4, 7, and 10 from panel a. *c*, Jaccard similarities between the learned representation of toluene and the activity patterns generated by plume-varying toluene stimuli across the 5 gamma cycles of each of the four sniffs depicted in panel b. In the present dataset, variance arising from plume dynamics is substantially more modest than that generated by impulse-noise based destructive interference. *d*, Ten sniffs of toluene drawn from randomly-selected timepoints within the dataset and also occluded with impulse noise (P = 0.4). *e*, Higher-resolution depictions of sniffs #1, 4, 7, and 10 from panel d. *f*, Jaccard similarities between the learned representation of toluene and the activity patterns generated by plume-varying, strongly occluded toluene stimuli across the 5 gamma cycles of each of the four sniffs depicted in panel e. *g*, Network activity evoked by presentation of plume-varying and occluded instances of each of the ten learned odors following one-shot learning. *Left panels*, spike raster plots over five gamma cycles; *ordinate* denotes MC index. *Right panels*, Jaccard similarities between the activity pattern generated by each occluded odorant stimulus and the learned representation of the corresponding odorant. The same network reliably recognized all ten odorants from plume-varying and substantially occluded examples (P = 0.4). *h*, The Jaccard similarity to toluene that was evoked by the occluded, plume-varying toluene stimulus increased over five successive gamma cycles until the stimulus was classified as toluene (similarity > 0.8). For clarity, only five odorants are depicted. *i*, Mean classification performance across all ten odorants, with plume dynamics, under increasing levels of sensory occlusion (100 impulse noise

instantiations per odorant per noise level). The abscissa denotes the level of impulse noise. *Green curve*, proportion of correct classifications by a network trained on all ten odorants.

**Figure 6**

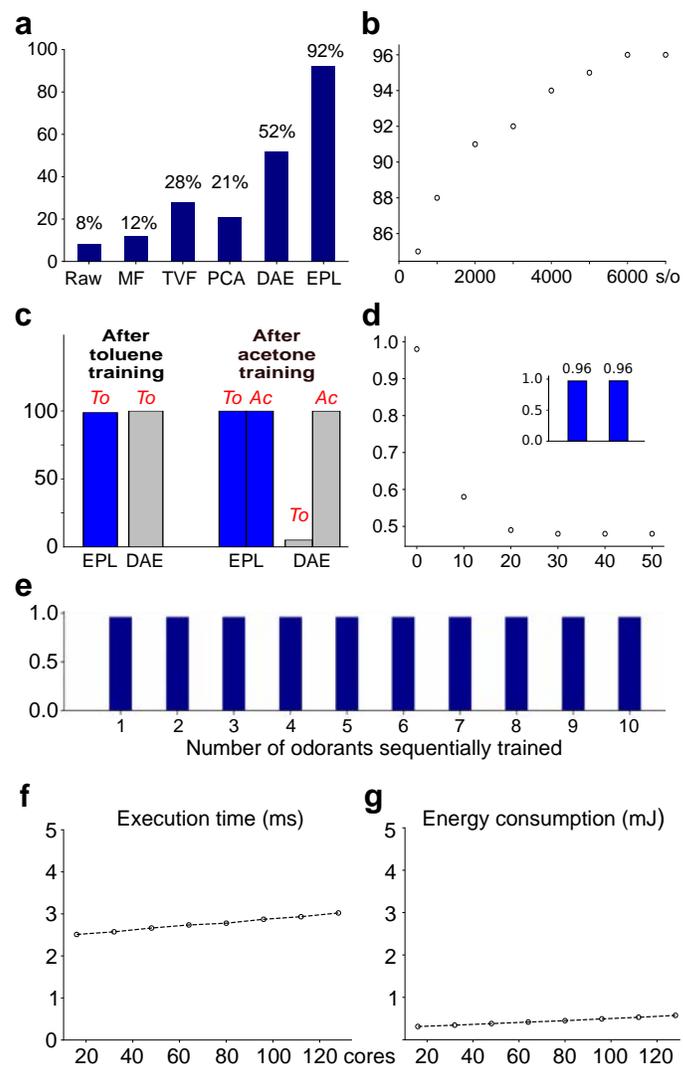

**Figure 6.** Performance evaluation. *a,* Classification performance of the EPL network in comparison to four other signal processing techniques. *Raw*, classification of unprocessed sensor signals. *MF*, median filter. *TVF*, total variation filter. *PCA*, principal components analysis. *DAE*, a seven-layer deep autoencoder. *EPL*, the neuromorphic EPL model. Each of the 10 odorants was presented with 100 independent instantiations of impulse noise, yielding 1000 total test samples. *b,* The performance of the DAE improved when it was explicitly trained to map a variety of occluded instances of each odor to a common representation. To achieve performance superior to the one-shot-trained EPL network, the DAE required 3000 occluded training samples per odorant. *Abscissa*, number of training samples per odorant (*s/o*). *Ordinate*, classification performance (%). *c,* Online learning. After training naïve EPL and DAE networks with toluene, both recognized toluene with 100% accuracy. After then training the same network with acetone, the DAE learned to recognize acetone with 100% accuracy, but was no longer able to recognize toluene (catastrophic forgetting). In contrast, the EPL network retained the ability to recognize toluene after subsequent training on acetone. *d,* Gradual loss of the toluene representation in the DAE during subsequent training with acetone. The ordinate denotes the similarity of the toluene-evoked activity pattern to the original toluene representation as a function of the number of training epochs for acetone. Values are the means of 100 test samples. *Inset*, Similarity between the toluene-evoked activity pattern and the original toluene representation in the EPL network before training with acetone (left) and after the completion of acetone training (right). *e,* Similarity between the toluene-evoked activity pattern and the original toluene representation as the EPL network is sequentially trained on all 10 odorants of the dataset. Values are the means of 100 test samples. *f,* The execution time to solution is not significantly affected as the EPL network size is expanded, reflecting the fine granularity of parallelism of the Loihi architecture. In the present implementation, one Loihi core corresponds to one OB column. **g,** The total energy consumed increases only modestly as the EPL network size is expanded.

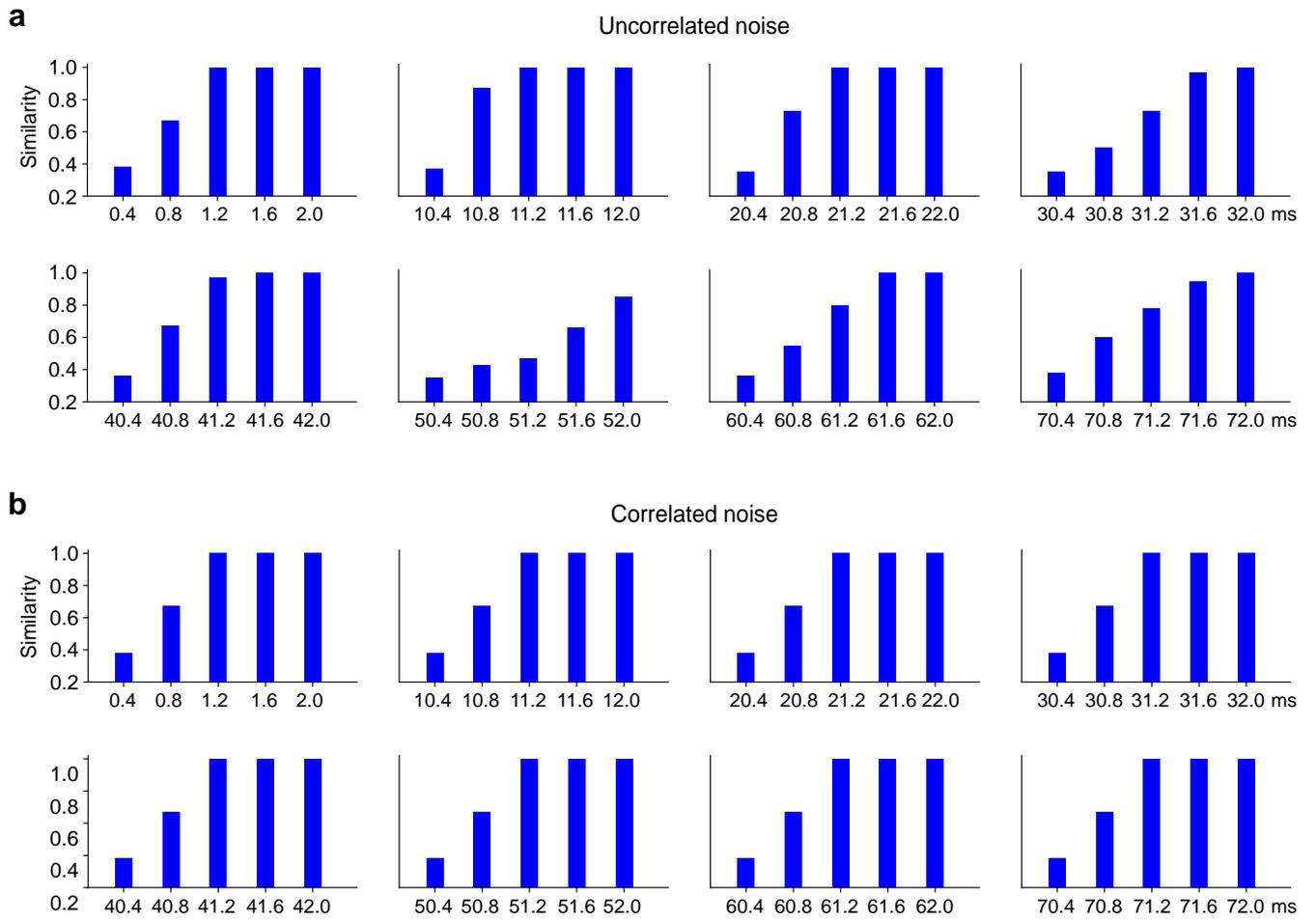

**Figure S1.** Illustration of "continuous" sampling by a trained network with impulse noise uncorrelated or correlated in time. *a,* Processing of eight immediately successive samples of a toluene plume from the Vergara et al. dataset [4], sampled at 100 Hz (10 ms per sample). Each sample was processed over five successive gamma cycles, requiring a total of 2 ms (see *Methods*). The instantiation of impulse noise ($P = 0.5$) was randomized for each sample. *b,* As in *a*, except that a single instantiation of impulse noise ($P = 0.5$) was maintained across all eight successive samples, modeling the continued presence of a single set of occluding inputs. The algorithm is indifferent to the presence or absence of these noise correlations over time.

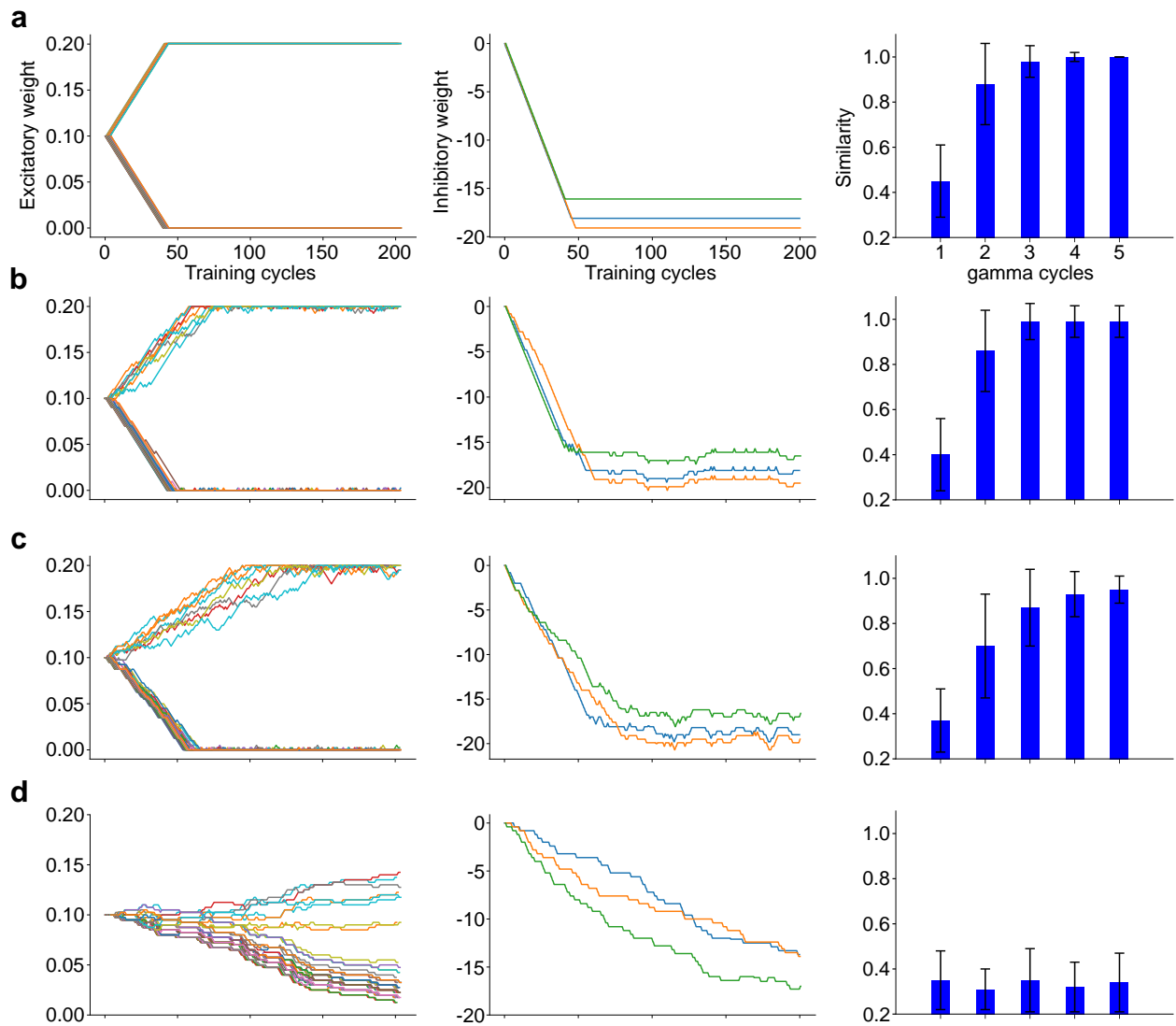

**Figure S2.** Few-shot training of toluene with destructive interference. *a*, No impulse noise during training. *Left panel*: Evolution of the excitatory weights of a set of MCs onto one GC over the course of training (learning rate of excitatory plasticity set to 0.005). Synapses are color coded. As training progresses, the weights of an odorant-specific set of synapses increase, while the weights of all other synapses decrease. *Center panel*: Evolution of the inhibitory weights of three GCs onto one MC over the course of training (learning rate of inhibitory plasticity set to 0.1). As training progresses, the weights converge to values that reflect the timing difference between pre- and post-synaptic spikes (see Methods for details). *Right panel*: After training, the network attracts test samples of toluene to the learned representation over the course of five gamma cycles. Graphs depict the similarity between test samples of toluene and the learned representation of toluene, averaged across 100 test samples. Impulse noise was randomly selected from the range [0.2-0.8] to generate each test sample. *b-c*, Same as *a*, but with impulse noise during training set to 0.2 and 0.4 respectively. Excitatory and inhibitory weights gradually converge to their respective values, despite the destructive interference. After training, the network accurately recalls the learned representation of toluene over the course of five gamma cycles. *d*, Same as *b-c*, but with impulse noise during training set to 0.6. Noise dominates the training process for this level of destructive interference, and excitatory and inhibitory weights do not converge to their correct values within 200 training sniffs. After training, the network is unable to recall the learned representation of toluene from the corrupted test samples. Results of panels *a-d* were generated using a software model of Loihi.